# Multi-label Classification for Fault Diagnosis of Rotating Electrical Machines

Adrienn Dineva*, Amir Mosavi*, Mate Gyimesi, and Istvan Vajda
Institute of Automation, Kalman Kando Faculty of Electrical Engineering, Obuda University, 1034 Budapest, Hungary

**Abstract**: Primary importance is devoted to Fault Detection and Diagnosis (FDI) of electrical machine and drive systems in modern industrial automation. The widespread use of Machine Learning techniques has made it possible to replace traditional motor fault detection techniques with more efficient solutions that are capable of early fault recognition by using large amounts of sensory data. However, the detection of concurrent failures is still a challenge in the presence of disturbing noises or when the multiple faults cause overlapping features. The contribution of this work is to propose a novel methodology using multi-label classification method for simultaneously diagnosing multiple faults and evaluating the fault severity under noisy conditions. Performance of various multi-label classification models are compared. Current and vibration signals are acquired under normal and fault conditions. The applicability of the proposed method is experimentally validated under diverse fault conditions such as unbalance and misalignment.

1. Introduction

Rotating electrical machines are responsible for converting a great amount of worldwide energy into mechanical energy [1-3]. Mobility, transportation, logistics, construction, production, agriculture, food, automation, and basically, any economical activities and industries directly or indirectly depend on rotating electrical machines [4-

6]. The rapidly evolving industries have suggested that we will be witnessing further increase this rate [7-12]. Furthermore, the increasing demand for the hybrid and electric vehicles, the rapid transition toward automated systems and micro and nano mechatronics devises, increasing interests for more efficient energy conversion systems, and emerging new robotics machines have been motivating further advancement in the rotating electrical machines [13-18].

One of the key factor of overall efficiency maximization covers the well-sized and high-efficient components [19-22]. Therefore, the reduction and prediction of faults occurring in electrical machines and drive systems such as electrical, thermal, mechanical faults of electrical machines are strongly suggested to be essential [23-29]. Classical solutions of fault diagnosis and identification (FDI) [30] are based on the complex mathematical models [25-29, 31], or dynamic models [32-37] of the processing system. Intelligent modernization has contributed to the widespread use of Machine Learning (ML) techniques in industrial applications [38-43]. As a result, the latest FDI systems demand more artificial intelligent solutions to incorporate multiple fault events or dynamically changing load profiles in case of incomplete or noisy measurements [44-49]. Commonly, the diagnosis and predictions is calculated through current signature analysis (MCSA) [50, 51], i.e., examining the output signals of the motor stator's current while running on a steady-state operating mood [52-56]. MCSA analyses the time-frequency decomposition of the current signals or by faults' frequencies in the frequency domain. MCSA works based on a single input source, and representing a simple, low-cost and non-invasive monitoring method [50, 57, 58]. An enhanced method of MCSA in case of multiphase electrical machines is called electrical signature analysis (ESA) [59].

Timely diagnosis of the complete rotating machinery system contributes to avoiding overpriced reparations and unexpected breakdowns. According to [60], the great majority of recent electric motor condition monitoring methods can be classified into three main categories. The first-class includes the detection of single faults by analyzing one or multiple parameters; the second class covers the detection of different faults with multiple parameters and processing techniques, and the last one contains the mixed techniques of various computing-intensive approaches to analyze *different electrical and mechanical*

*parameters in order to detect multiple faults [61-64]*. In contrast to conventional signal processing based fault detection techniques [65], recently a few attempts are made for the application of intelligent algorithms [66, 67] including new approaches to fault detection and isolation (FDI) [68] based on fuzzy logic, decision trees, neural networks, and further machine learning techniques [69-73]. However, most of them rely on the measurement and processing of vibration signals, which require at least one vibration sensor, which demands extra costs for its proper installation and maintenance [74-77]. In addition, a technician needs knowledge and a good amount of experience to correctly use such sensors [78-84]. However, the ESA reported to be able to reveal a large number of relations between the machine parameters [85-88]. Therefore, the ML techniques are highly suitable to support the processing of such extracted information.

There are some intelligent methods which have evolved in recent years with the purpose of improving fault diagnosis methods of electrical motor and drive systems associated with various fault events on the basis of current or vibration signature analysis. For instance, in [89] it has been shown that motor current signature analysis based Support Vector Machine classifier achieves better accuracy in case of gearbox multiple fault detection than decision trees. In [90] the authors state that hybridization of machine learning methods, e.g., adaptive neuro-fuzzy inference system (ANFIS) in combination with Classification and Regression Tree (CART) has the potential for fault diagnosis of induction motors. The proposed method also employs vibration and current signals and achieves nice performance, however, the method lacks the flexibility to easily fit with specific types of machines. Random Forest Classifier was developed for multi-class bearing faults in [91]. This work also employs input features extracted only from vibration signals.

An number of studies report the advantages of the wavelet transform in signal detection and fault feature identification (see, e.g. [92-96]). The decision tree's capability, with combination the well-studied wavelet technique also provides a possible tool for fault detection [97]. In [97], a large number of possible wavelets are analyzed in order to find the best match. The selection of a suitable wavelet function is still a challenge for the users for specific applications. The investigation of mechanical fault signatures of misalignment and unbalance is carried out in [97], where the authors found that a multilayer perceptron model only with three layers is able to classify the faults. High accuracy is reported that is achieved by training the network on a large dataset. In [98] a Convolutional Neural Network (CNN) is successfully applied for fault classification. This approach relies on the S-transform of vibration signals into images displaying the time-frequency patterns in which the pattern recognition is performed. We can clearly observe that the common pattern recognition methods are relying on the assumption of a single fault scenario and only a limited number of work proposes flexible and comprehensive solutions for multiple fault detection. In the presence of multiple faults, the single fault recognition techniques' performance may be degraded. In addition, the other machines operating in the motor's environment or the coupled subsystems, etc., may introduce further noise components in the measured signal. As a result, it may occur that one of the fault or noise component obscure a particular fault feature and makes it impossible to recognize it. Its further consequence is that the fault isolation operations may become rather difficult.

Based on the above briefly summarized antecedents and state-of-the-art applications this research on the development of new condition monitoring and diagnostic methodology for electrical machines and drives are focused into the applicability of multi-label classification machine learning approaches for multiple fault detection under noisy conditions and also the simultaneous determination of fault severity. The contribution of this work is to propose a novel methodology using multi-label classification method for simultaneously diagnosing multiple faults and evaluating the fault severity under noisy conditions. Performance of various multi-label classification models are compared. Current and vibration signals are acquired under normal and fault conditions.

The rest of this paper organizes as follows: in Section II. we briefly introduce the latest multi-label classification methods, and we derive a new methodology for multi-fault diagnosis and severity assessment for rotating electrical machines and drive systems. Section III is devoted to the description of the experimental setup as well as the description of feature extraction and dataset preparation. Section IV presents the results of our investigations on the multi-label classification methods. After, in Section V. we draw some pertinent conclusions.

**2. Multiple Fault Classification and Fault Severity Determination**

*2.1. Brief Introduction of Multi-label Classification Methods*

Modern industry requires data mining algorithms that are able to efficiently cope with the growing amount of information and large datasets. Specialized processing tasks of various practical application fields deal with common characteristics of the stored data that can be assigned to multiple categories [99, 100]. Therefore, multi-label classification algorithms have gained increasing interest in recent years. Specialized techniques for learning such type of data is still in the focus of researches which have the capability of predicting a set of relevant labels for new species. Currently, three main groups of newly developed multi-label classification methods are proposed in the literature, namely the data transformation methods, adaptation methods and ensemble of classifiers [101, 102]. Early solutions for multi-label classification methods cover the data transformation techniques. The concept is to turn the original multi-label set into binary sets or multi-class sets that adequately can be processed with the classical algorithms. Besides binarization with the widely applied binary relevance technique, also the voting methods and divide-and-conquer approaches are also applied for accomplish multi-label transformation [100, 103, 104]. Such separated sets are learned by single-label classifiers, such as decision trees. This group also includes the Classifier Chains that is similar to the binary relevance technique, but it performs the binarization in consecutive classifiers. Since, the models' order has importance [105, 106]. The output of one response variable for a sub-classifier is used as an additional feature in the next sub-classifier. Optimal order of the classification models can be enhanced by the Naïve Bayes method.

Adaptation methods are based on the adaptation of conventional classification methods to multi-label versions without problem transformation. The adaptation methods are extensions of well-founded automatic classifications algorithms. For instance, the support vector machine (SVM) classifiers [107] or the k-nearest neighbors (KNN) classifiers [108, 109] are able to predict binary or multiclass outputs simultaneously [110, 111]. KNN is a non-parametric method used for classification and regression [112, 113]. In KNN, an object is classified through a plurality vote of its neighbors while the object assigned to the class of the most common k nearest neighbors. K is a positive and small number, to be able to reduce the number of calculations, and thus the process duration. Similarly, the neural networks and decision trees have such abilities, but the adaptation of the algorithm may become difficult. Ensemble classifiers are also held notable interest. The classification ensemble approach is based on the aggregation of the outputs of a number of the individual classifier by weighted or unweighted averaging [114]. According to the 'no-free-lunch' theory [115, 116], it is assumed that the weaker classifiers with different bias can achieve better performance than the better ones [114, 117, 118].

*2.2. The Scheme of the Proposed Method*

We mainly focus on diagnostics based on Electrical Signature Analysis, but we also utilize traditional vibration data. The analysis starts with the elimination of the components that do not contribute to the system from the acquired stator phase current signals. After, the filtered signal is used for extracting the fault features affecting the machine health. Then, the trained multi-label classifier receives the new feature vectors and performs the multi-label prediction. The flowchart of the method is depicted in Fig. 1. below.

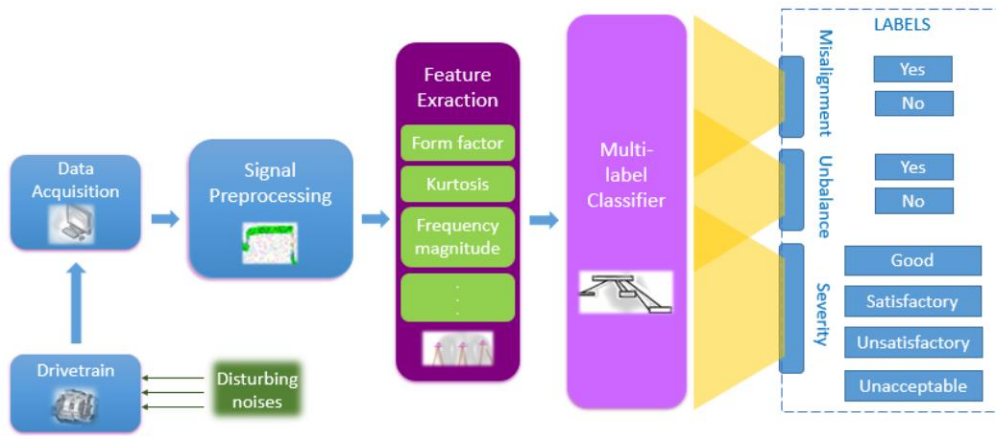

Figure 1. Flow chart of the proposed multi-label fault diagnosis system

For both of the misalignment and unbalance faults the two targets can be treated as logical labels, i.e. each output is a binary value, indicating whether a label is associated with the fault or not.

Table I. Vibration severity per ISO 10816.

| vibration velocity [mm/s] | Class I. small machines | Class II. medium machines | Class III. large rigid foundation | Class IV. large soft foundation |
|---|---|---|---|---|
| 0.28 | | | | |
| 0.45 | | GOOD | | |
| 0.71 | | | | |
| 1.12 | | | | |
| 1.80 | | | | |
| 2.80 | | SATISFACTORY | | |
| 4.50 | | | | |
| 7.71 | | UNSATISFACTORY | | |
| 11.20 | | | | |
| 18.00 | | | | |
| 28.00 | | UNACCEPTABLE | | |
| 45.90 | | | | |

For evaluating the fault severity, we can inject additional classes or alternatively we can generate a second classifier which is executed parallel. Acquired vibration data is sufficient for proper evaluation. The fault severity can be determined easily according to the ISO 2372 Standard [119, 120] for vibration severity of machines operating between 600 to 12.000 rpm range. The severity classes are shown in Table 1.

**3. System Description and Experiment**

*3.1. System Description*

The outlined method requires precise measurements and appropriate dataset. The theoretical considerations and their usability are validated by simulation investigations by using the collected data on the test bench. The Laboratory of Electrical Machines of the Institute of Automation provides equipment for motor diagnostics and expert system development that allows the integration of measurement, computing, and communication. The laboratory has a *wind power simulation system,* where the wind power is simulated by an inverter-driven cage induction machine (Leroy Somer 3fFLSE225M-TC; No28221L12001/2012; IP55IK08; P=30 kW; n=985 1/min; U=230/400 V; f=50 Hz; cos φ=0,82). Three different types of motors are available on the test bench: one brushless synchronous machine of 40 kW, one double-fed asynchronous machine and one permanent magnet synchronous machine (PMSM) (Leroy Somer 3fLSRPM200L-T; No728333K12001/2012; IP55IK08; P=40 kW; n=1500 1/min; U=400 V; f=100 Hz; I=83 A; 95,2 %; Imax/In=145 %) on which the measurements and tests are carried out (see, Fig. 2.).

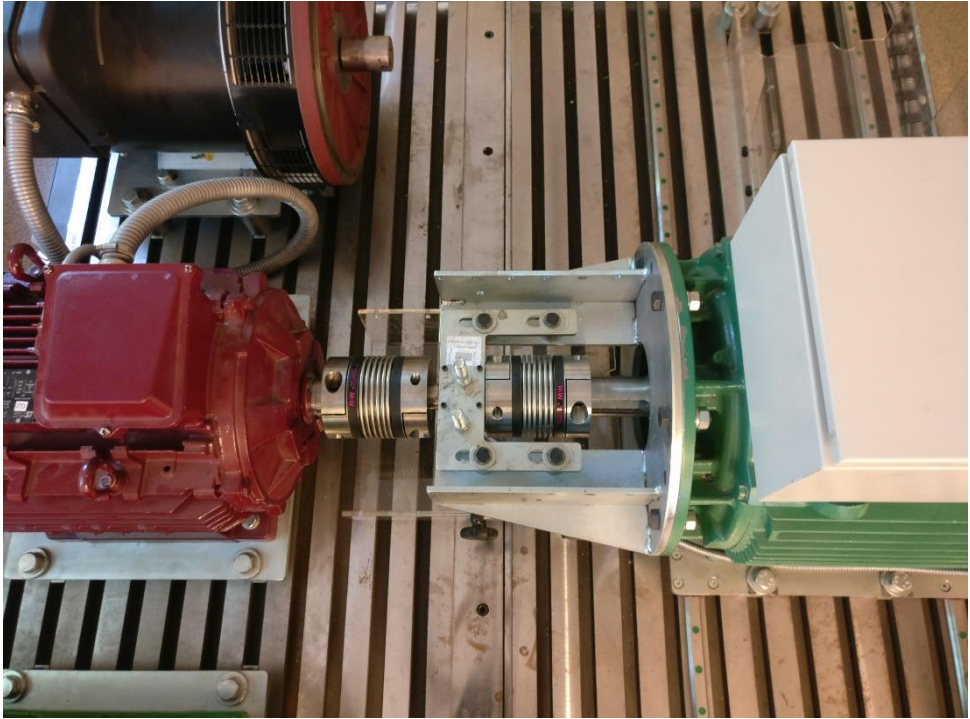

Figure 2. Bellows shaft coupling connects the generator (left) and the motor (right).

The system is capable of simulating fault events. Misalignment and unbalance is introduced in the system and a simple NI data acquisition system for data collection purposes is installed. The laboratory has the latest NI LabVIEW Software. The data acquisition system consists of the National Instruments PCI 6013 B-Series 16-Analog-Input multifunction DAQ board and the SC-2345 signal conditioning connector block with various modules and sensors connected to the test bench. The representation of the data acquisition system through a visualization of block schematic illustraion is available in Fig. 3.

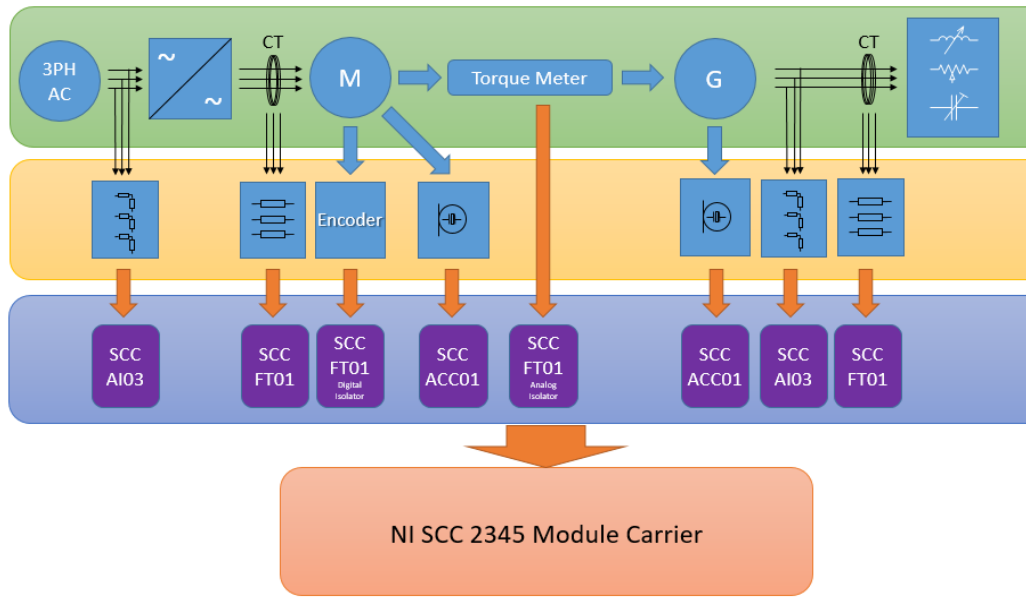

Figure 3. Block scheme of the data acquisition system.

The measuring and controlling LabVIEW software collaborates with the above-presented devices and additional hardware components and conducts the measurements. The measurement data is processed simultaneously on each channel which takes a relatively short time. The LabVIEW software is developed in order to inspect the proper configuration of the test bench. The software displays the amplitude-time signals of the three-phase stator current and vibration signals of the motor and the generator in real-time. The second tab of the software views the Park Vector's spectrums calculated from the current signals. Frequency-amplitude diagrams can be used to predict machine errors during measurement. The third part of the program saves the measurement results in the Excel .csv format, so the measured values can be evaluated and processed later.

*3.2. Feature Extraction and Dataset Preparation for Training*

Stator current signals and vibration data picked for fault-free no load and for 40% of full load at the speed of 1500 rpm conditions with fs=10000 sampling rate and t=2 seconds duration in the presence of both mechanical faults. Unbalance results in high-frequency amplitudes at frequencies at once the rotational speed. The typical fault features of misalignment are dominant frequencies at one or two times the rotational speed depending on the degree of angular misalignment and the type of the couplings. Spectral images also display sub-harmonic multiplies of 1/2xRPM. However, the different speeds, loads, motor parameters and operational setups, etc. may affect the fault frequencies. Differences between faulty and healthy conditions can immediately be observed from vibration spectra during the measurement (see, Figure 4.).

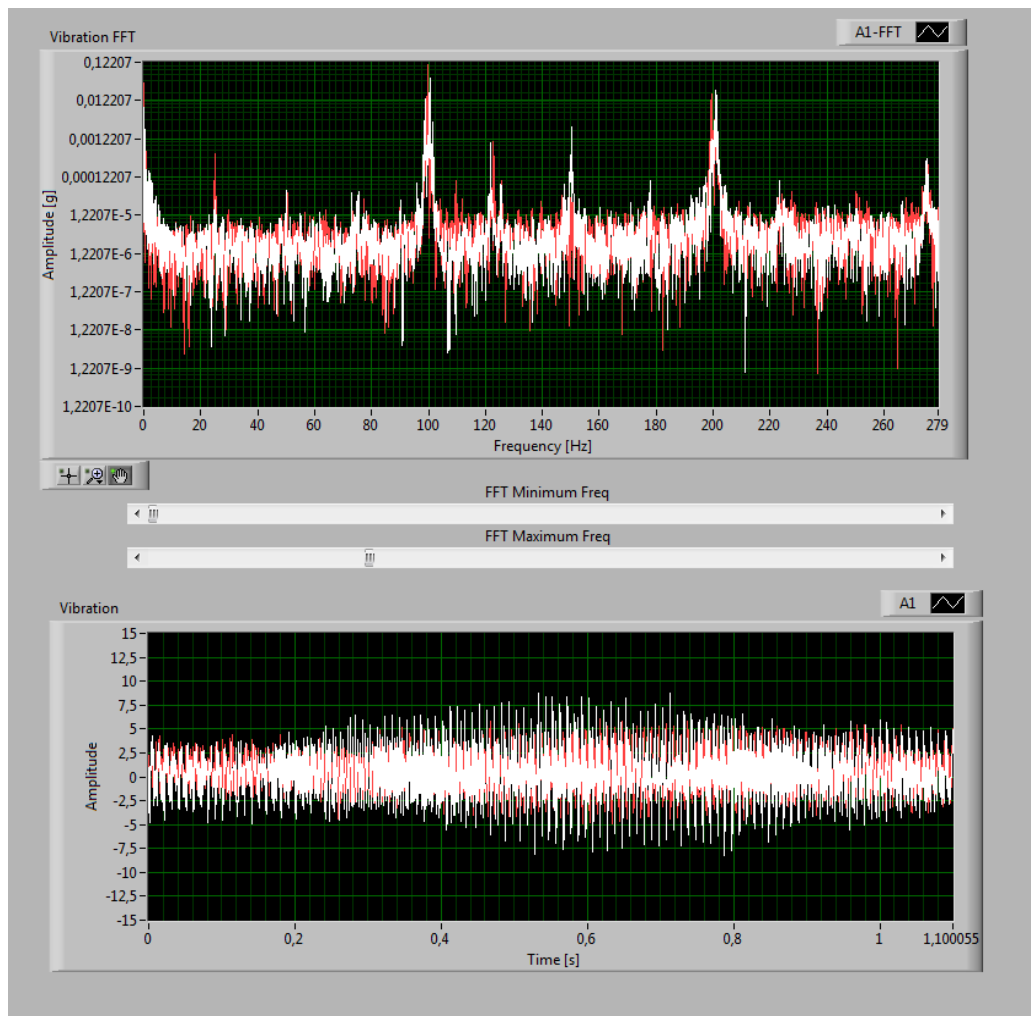

Figure 4. Vibration FFT spectrum (upper chart) and time-domain signal (lower chart) during measurement displayed by the LabVIEW software. A1 (white line) denotes the generator's vibration signal and A2 (red line) stands for the motor's vibration signal.

It is clearly visible in the spectrum that the machine vibrates strongly around 25 Hz, which indicates a shaft misalignment. From the generator current spectrum, a 50 Hz component appeared in the spectra, while in the motor current spectrum a 75 Hz component. The reason is that, the generator is a 4-pole machine, while the motor has 6 poles. This indicates exactly that the machine vibrates at 25 Hz.

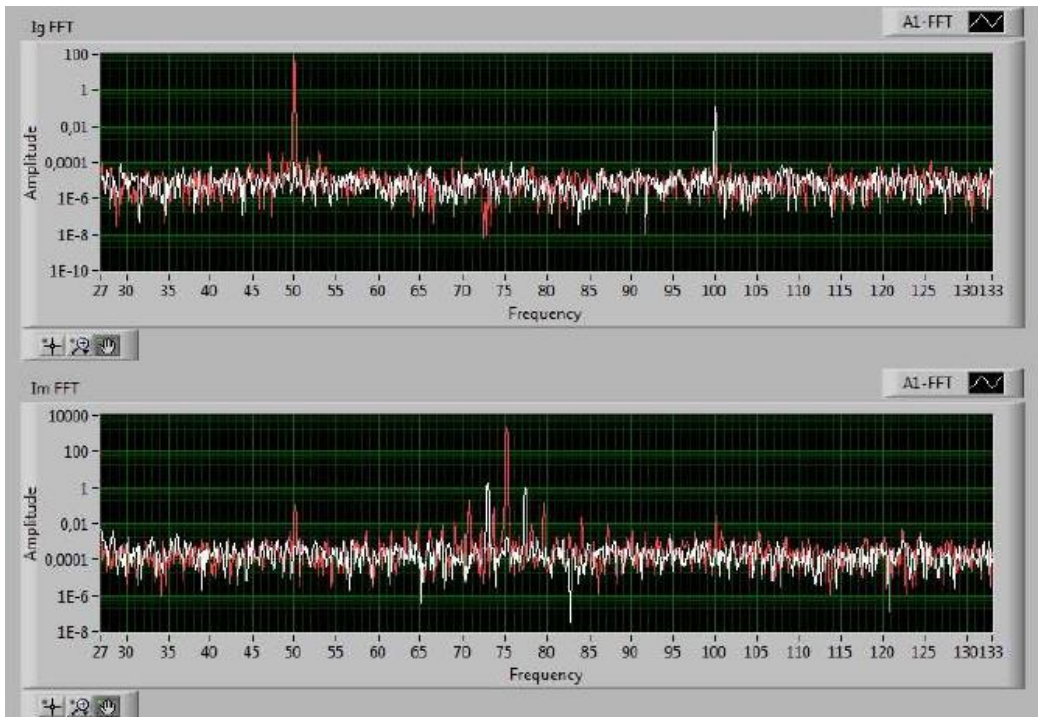

Figure 5. FFT spectra of the generator current (upper chart) and motor current (lower chart).

The unbalancing load is placed at the generator's side, so the flexible bellows coupling may attenuate the fault signature in its spectra. Therefore, for fault extraction, we have applied the Thomson multitaper spectral estimates that are proven to be efficient in case of weaker signals [121] and combines the beneficial properties of high resolution and low variance. Figure 6. displays the Thomson multitaper spectrum of the generator's current signal in which we can observe how the distinguished magnitudes are emphasized. This allows a fast automatic extraction of the peaks.

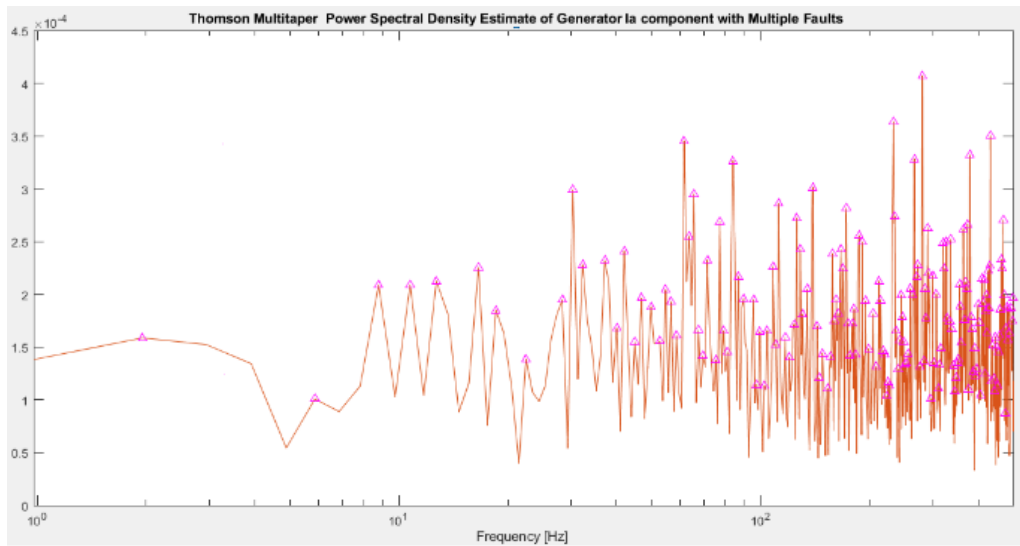

Figure 6. Multitaper power spectral density estimation of generator current signal for fault frequency magnitude detection.

The features carrying most of the information of interest are mainly in the frequency domain. However we have inspected a few time-domain features also. The feature vector composed from the magnitudes of the distinguished sideband frequencies and RMS variance frequencies from the spectra for all three phases of both machines. The time-domain features cover the form factor, kurtosis and the entropy deviation from the fault-free sample. Fault feature vector includes also the distance calculated between the observed signature and each pure signature associated with the identified fault [122]. The dataset is built by synthetically adding a random number of ten to twenty of external disturbing frequency components for the faulty and fault free signals. Fifteen different samples are generated with random contaminating frequencies for all of the fault-free, multiple-fault and single fault (unbalance or misalignment) cases and additionally the data collected by the accelerometers at both sides for the severity evaluation. The training set consists of 64 feature vectors including the original ones and each of them containing 27 features.

Table 2. Structure of the dataset.

| id | Imkf | Imbkf | Imckf | ... | i | ... | isUnbalance | isMisalignm | Severity |
|----|------|-------|-------|-----|---|-----|-------------|-------------|----------|
| 0 | 0.687150 | 0.028560 | 0.123693 | ... | 0.358956 | | 1 | 1 | 'Good' |
| 1 | 0.019699 | 0.388238 | 0.07809 | ... | 0.900697 | | 1 | 0 | 'Good' |
| | | | | . | | | | | |
| | | | | . | | | | | |
| | | | | . | | | | | |

In our qualitative models '*isUnbalance*' and '*isMisalignment*' (Table 2.) are the fault label columns which consists of values 0 (No) and 1 (Yes) corresponding to the presence of the symptom of a fault in a given sample. For the fault severity classifier, we have specified the labels according to the previously presented ISO standard [25] as '*Good*', '*Satisfactory*', '*Unsatisfactory*', '*Unacceptable*'. The dataset is divided into a training set 80% of the vectors to train the classifier and a testing set of 20 % for testing the accuracy of the classifier on new data. When most of the samples have the same labels in the training set, the classifier would achieve a high accuracy value. In order to avoid such a bias half of the dataset composed of samples including the fault features. Further metrics are applied for the proper evaluation of the performance.

## 4. Results

The experiments with three different algorithms are performed by applying the Python Scikit-multi-learn library which offers various classification approaches that are suitable for predicting simultaneously multiple outputs. At first, a binarized Decision Tree has been implemented for predicting the labels. As a criterion, which defines the function to measure the quality of a split both criteria for Gini Index and entropy for Information Gain were used. We have modified also the maximum of tree depth in order to compare the performance using various maximum depths of the trees. Various tree models were built also with including and excluding the error attribute values. According to initial tests we have found that by applying the error feature attributes in the training does not result in significant improvement in the accuracy. The best accuracy score we have achieved is 0.7333. The accuracy score is not the most representative metric in case of multi-label classification. Therefore, further metrics are applied. The results are collected in Table 3.

Table 3. Performance evaluation of the tested models.

|  |  | precision | recall | f1-score |
|---|---|---|---|---|
| **Binarized Decision Tree** | 0 | 0.79 | 0.88 | 0.83 |
|  | 1 | 0.85 | 0.94 | 0.89 |
| **Classifier Chain** | 0 | 0.79 | 0.88 | 0.83 |
|  | 1 | 0.78 | 1.00 | 0.88 |
| **KNN** | 0 | 0.76 | 0.76 | 0.76 |
|  | 1 | 0.82 | 1.00 | 0.90 |

The precision shows also the accuracy of the model by the ration of total predicted positive and the number of true positive ones. Recall or sensitivity is calculated by the number of true positives divided by the number of true positives plus the number of false negatives. The F1 score covers the weighted average of the sensitivity and precision values that is suitable to characterize the test performance. After, a Classifier Chain multi-label method using Gaussian Naïve Bayes approach has been tested. We found it to be more efficient. Its performance can be seen in Table 3. The Classifier Chain's score accuracy resulted in 0.8333. Subsequently, we have tested the multi-label KNN method whose best percentage accuracy resulted in 0.7 after testing the algorithm with various k values.

It can be seen from Table 3. that the classification performance has been the most enhanced by using Classifier Chains and KNN. The accuracy may be further improved by training a model with a larger training set and further tuning the algorithms. Its excellent pattern recognition capabilities can be effectively utilized for the fault classification of electrical machines in the presence of disturbing noises. The prediction performance of the parallel severity classification tree resulted in 99% accuracy because most of the vibration data is labeled '*good*'.

## 5. Discussions

In modern industrial systems, complexity is increasing as multi-sensor network systems are expanding towards largescale systems. The intelligent solutions of Electrical Signature Analysis allow simplifying the fault identification processes because *does not require a large number of sensors* that results in remarkable *cost reduction* and *support sensorless and largescale technologies*. Furthermore, a well-developed theoretical and practical methodology could serve as a basis for a *reliable remote –diagnostics* tasks of electrical machines that are especially important diagnostics carried out in hazardous environment (for instance, in nuclear plants). The development of appropriate diagnostic-prediction methods which are capable of reliably and timely evaluating the health status of the system on the basis of representative parameters acquired directly or indirectly is an important part of the modern electric drivetrain monitoring system. The state-of-health of a certain dynamic system and its possible initial failures can be addressed by a number of approaches published in the literature. The strengths and weaknesses of conventional fault diagnosis methods have already been proven. In recent years, diagnostics research has focused on the prediction algorithms that can identify the fault features of progressive malfunctions. As such algorithms are highly technology-dependent, it is important to define the method as a function of a large number of parameters of the dynamic system for fitting to the system under consideration, taking into account boundary conditions. Data-driven technologies aim such difficulties.

Fault classification methods have already established for the investigation of the relations between the symptom and the fault feature. It is clear that binary relationships can be easily represented with such systems. Early versions of these algorithms are used rather for visualizing diagnostic reasoning. However, practical engineering problems demanded the development of new automatic methods that are able to deal with multiple fault scenarios and noisy or uncertain fault features. A possible solution is the extension of the most commonly applied classification methods. Recently binary trees, Chain Classifiers, Neural Networks, etc. and other multi-label and multi-class techniques are introduced in the literature. However, these algorithms need more development and improvement especially in terms of robustness. In addition, the application of Machine Learning techniques on sensory data is still not well-established and requires the synergy of classical signal processing and intelligent data analysis. As a conclusion from all cases studied, the analysis of motor current signatures in combination with the machine learning-based processing and evaluation algorithms is a viable methodology for fault detection and prediction in electrical machine and drive systems that has the following *additional advantages*; the measurement or monitoring *does not interfere with the production process*; it is *well suited for any machine* with speed, performance and power; it is *universal* because it is able to detect the quality of electricity supply, disturbances, faults in a motor control, starter and control- devices, the efficiency of the electric machine and its characteristics, errors in the electrical machine, plant-wide machine monitoring, and its operating costs (energy consumption, repair, maintenance) meet the economy's expectations. The transfer of recent Machine Learning approaches to practical fault diagnosis problems of rotating electrical machines and drive systems may help to facilitate smart industrialization and intelligent modernization.

## 6. Conclusions

This paper proposes a novel methodology using multi-label classification method for simultaneously diagnosing multiple faults and evaluating the fault severity under noisy conditions. Furthermore, the performance of various multi-label classification models are compared. Current and vibration signals are acquired under normal and fault conditions. The applicability of the proposed method is experimentally validated under diverse fault conditions such as unbalance and misalignment. The prediction performance of the parallel severity classification tree resulted in 99% accuracy and most of the vibration data is labeled '*good*'.

**Author Contributions:** Principal investigator, lead author, writing, machine learning expertise, electric machines expertise, A.D.; data collection, data curation, data analysis, technical expertise, laboratory expertise, data acquisition system implementation, LabVIEW programming, M.G.; machine learning modeling expertise, writing, revision support, A.M.; electric machines expertise, conceptualization, supervision, resources, software, controlling and verifying the results, I.V.

**Funding:** This work has been supported by the project GINOP-2.3.4-15-2016-00003.

**Conflicts of Interest:** The authors declare no conflict of interest